\documentclass[10pt, a4paper]{article}
\usepackage{lrec2022} 
\usepackage{multibib}
\newcites{languageresource}{Language Resources}
\usepackage{graphicx}
\usepackage{tabularx}
\usepackage{soul}
\usepackage{titlesec}
\titleformat{\section}{\normalfont\large\bfseries\center}{\thesection.}{1em}{}
\titleformat{\subsection}{\normalfont\SmallTitleFont\bfseries\raggedright}{\thesubsection.}{1em}{}
\titleformat{\subsubsection}{\normalfont\normalsize\bfseries\raggedright}{\thesubsubsection.}{1em}{}
\renewcommand\thesection{\arabic{section}}
\renewcommand\thesubsection{\thesection.\arabic{subsection}}
\renewcommand\thesubsubsection{\thesubsection.\arabic{subsubsection}}

\usepackage[utf8]{inputenc}
\usepackage{booktabs}

\usepackage{hyperref}
\usepackage{xstring}
\usepackage{amsmath}
\usepackage{url}

\usepackage{CJKutf8}

\usepackage{color}

\title{Construction of a Quality Estimation Dataset for Automatic Evaluation of Japanese Grammatical Error Correction}

\name{\large 
$\text{Daisuke Suzuki}^\text{1}$, $\text{Yujin Takahashi}^\text{1}$, $\text{Ikumi Yamashita}^\text{1}$, $\text{Taichi Aida}^\text{1}$, 
$\text{Tosho Hirasawa}^\text{1}$,\\
\large \textbf{
$\text{Michitaka Nakatsuji}^\text{1}$, 
$\text{Masato Mita}^\text{2,1}$,
$\text{Mamoru Komachi}^\text{1}$}}

\address{
$\text{Tokyo Metropolitan University}^\text{1}$ \\
$\text{RIKEN}^\text{2}$\\
         \{suzuki-daisuke3@ed., takahashi-yujin@ed., yamashita-ikumi@ed., aida-taichi@ed.,
         hirasawa-tosho@ed., \\
         nakatsuji-michitaka@ed.,\}tmu.ac.jp, 
         masato.mita@riken.jp, komachi@tmu.ac.jp\\}

\abstract{
In grammatical error correction (GEC), automatic evaluation is an important factor for research and development of GEC systems.
Previous studies on automatic evaluation have demonstrated that quality estimation models built from datasets with manual evaluation can achieve high performance in automatic evaluation of English GEC without using reference sentences.
However, quality estimation models have not yet been studied in Japanese, because there are no datasets for constructing quality estimation models.
Therefore, in this study, we created a quality estimation dataset with manual evaluation to build an automatic evaluation model for Japanese GEC.
Moreover, we conducted a meta-evaluation to verify the dataset's usefulness in building the Japanese quality estimation model. 
\\ \newline \Keywords{corpus construction, automatic evaluation, quality estimation, Japanese grammatical error correction} }

\begin{document}

\maketitleabstract

\section{Introduction}
Grammatical error correction (GEC) is the task of correcting different kinds of errors in text such as spelling, punctuation, grammatical, and word choice errors.
Automatic evaluation is considered as an important factor for research and development of GEC systems.

There are two types of automatic evaluation systems for GEC: reference-based and reference-less.
The former has been used extensively in the GEC community,
while the latter has gained much attention recently.

One of the problems of reference-based methods is the diversity of references.
In a study by \citelanguageresource{bryant-ng-2015-far}, a dataset with corrections was created by 10 annotators on the CoNLL-2014 test set \citelanguageresource{ng-etal-2014-conll}.
They found that the number of error types corrected by each annotator varied significantly, indicating the existence of a variety of corrections for a sentences containing error.
However, with a limited number of reference sentences, it is difficult to cover such a variety of corrected sentences in GEC.
So reference-based automatic evaluation methods may give unreasonably low scores to system outputs that are validly corrected but unseen in the references.

On the other hand, reference-less automatic evaluation methods (quality estimation models) do not have this problem because it estimates the quality of the system's output without requiring gold-standard references.
\newcite{asano-etal-2017-reference} and \newcite{yoshimura-etal-2020-reference} built a quality estimation model using a dataset with manual evaluations,\footnote{\newcite{asano-etal-2017-reference} have built several quality estimation models, and one of them uses a dataset of manual evaluation values.} and achieves a higher correlation with manual evaluation than with reference-based methods.

However, there is no quality estimation dataset for Japanese, although Japanese is a language with a large number of language learners.
This means no reference-less automatic evaluation method has been proposed for Japanese GEC.
In the NAIST Lang-8 Learner Corpora \citelanguageresource{mizumoto-etal-2011-mining}, one of the largest corpora of language learners, out of its composition of 580,549 essays written by language learners, 185,991 (which is the second-largest number after English) are written in Japanese.
This means that the study on quality estimation models for Japanese GEC can be large impact to GEC community.

Thereby, in this study, we created a quality estimation dataset with manual evaluation to build an automatic evaluation model for Japanese GEC.
The dataset consisted of three components: source text, corrected texts, and manually evaluated scores.
The source text comes from the Lang-8 corpus, and the corrected texts consisted of the output of four diverse GEC systems.
We built a quality estimation model by finetuning a pre-trained sentence encoder (namely BERT \cite{devlin-etal-2019-bert}) on the created dataset, and calculated the correlation with the manual evaluation values.
We also calculated correlations with manual evaluations for reference-based automatic evaluation methods and compared them with the correlations of the quality estimation model to meta-evaluate the quality estimation model built using this dataset.

The main contributions of this study are as follows.
\begin{itemize}
    \item{To build a quality estimation model for Japanese GEC, we created a dataset of multiple system outputs annotated with human evaluation.}
    \item{We demonstrated that the quality estimation model for Japanese GEC performs better than the reference-based automatic evaluation by building a quality estimation model using the created data set and evaluating its performance.}
 \end{itemize}


\section{Related Work}

\subsection{Evaluation Method}
\paragraph{Reference-based methods.}
In the early stage of English GEC, system outputs were evaluated by calculating the match rate, recall rate, and F-score per word \cite{dale-kilgarriff-2011-helping}.
After the success of shared tasks in GEC, Max Match (M$^2$), which calculates the match rate, recall rate, and F$_{0.5}$ score per phrase, has been widely adopted in the NLP community.
However, despite the prevalence of the M$^2$ scorer, \newcite{felice-briscoe-2015-towards} proposed I-measure, which gives a score of $-1$ to $1$, while other evaluation methods give a score of $0$ to $1$, so that the bad corrections will receive low scores while good corrections will gain high scores.
Further, \newcite{napoles-etal-2015-ground} proposed GLEU, which is a modified version of BLEU \cite{papineni-etal-2002-bleu} for evaluating GEC.
BLEU evaluates by comparing the translated sentence with the reference sentence, while GLEU evaluates by comparing three sentences: the source sentence, the corrected sentence, and the reference sentence.
Among the methods that used reference sentences, GLEU had the highest correlation with manual evaluation.

As for Japanese GEC,
\newcite{mizumoto-etal-2011-mining} conducted an automatic evaluation using BLEU to compare the grammaticality of the language learner's written sentences and the reference sentences.
More recently, \newcite{koyama-etal-2020-construction} conducted an  automatic evaluation using GLEU with an evaluation corpus for Japanese GEC created in their research.

\paragraph{Reference-less methods.}
Reference-less automatic evaluation methods, on the other hand, have been proposed recently.
\newcite{napoles-etal-2016-theres} evaluated a GEC system using the number of errors detected by the grammatical error detection system and showed that it performed as well as GLEU.
\newcite{asano-etal-2017-reference} proposed a method for evaluating correction sentences in terms of grammaticality, fluency, and meaning preservation, using logistic regression models for grammaticality, RNN language models for fluency, and METEOR \cite{denkowski-lavie-2014-meteor} for meaning preservation.
\newcite{yoshimura-etal-2020-reference} proposed a method to optimize the three evaluation measures for manual evaluation as an extension of \newcite{asano-etal-2017-reference}.
The method builds a quality estimation model by finetuning a pre-trained sentence encoder (BERT) on the manual evaluation of each measure and shows better correlation with the manual evaluation than \newcite{asano-etal-2017-reference}.

However, reference-less automatic evaluation has not been applied to Japanese GEC, while reference-based automatic evaluation is commonly used  \cite{mizumoto-etal-2011-mining,koyama-etal-2020-construction}. 
Therefore, in order to adapt a reference-less automatic evaluation method for the Japanese GEC, we created a dataset with manual evaluation values for building a quality estimation model.

\subsection{Dataset with Human Evaluation}
We introduce existing datasets with manual evaluation of corrected sentences as related work.

The GUG dataset \citelanguageresource{heilman-etal-2014-predicting} is a dataset consists of 3,129 sentences randomly sampled from English language learners' essays. The dataset contains five grammaticality ratings for sentences written by a single English learner, one rating by a linguistically trained annotator and five ratings by crowdsourced annotators.
The GUG dataset was created to measure the performance of automatic evaluation methods, and it has also been used as training data for quality estimation models.
In this study, the evaluation was done on a 5-point Likert scale, following the GUG data set.

\citelanguageresource{grundkiewicz-etal-2015-human} assigned human ratings to the output of 12 grammatical error correction systems that participated in the CoNLL2014 Shared Task on English GEC \citelanguageresource{ng-etal-2014-conll}.
In their dataset, the human ratings are annotated relative to the source text by ranking multiple corrections.

In this study, we created a dataset of manual evaluation as the training data for a quality estimation model for Japanese GEC.
Because a dataset with manual evaluation has not yet been created for Japanese GEC, the dataset created in this study will be useful for evaluating the performance of automatic evaluation methods for Japanese GEC.

\section{Construction of QE Dataset}
To construct the dataset, we first created pairs of error-containing and corrected sentences.
To create sentence pairs, we used four diverse GEC systems (Sec. \ref{Grammatical Error Correction System}) to generate correction sentences for two corpora of Japanese learners (Sec. \ref{Japanese Learner Corpus}).
We then manually evaluated the pairs of error-containing and corrected sentences (Sec. \ref{Annotation}).

\subsection{Grammatical Error Correction System}
\label{Grammatical Error Correction System}
We used the NAIST Lang-8 Learner Corpora \citelanguageresource{mizumoto-etal-2011-mining} for the training data. 
It was created by collecting essays written by language learners and their correction logs from 2007 to 2011 from Lang-8, a mutually corrected social media platform for language learners.
It contains the language learners' sentences and their corrections, essay IDs, user IDs, learning language tags, and native language tags in the JSON format.
The number of essays with the Japanese learning language tag is 185,991, and the number of sentence pairs of learners' sentences and corrections is approximately 1.4 million.

We employed four GEC systems to output four correction sentences for one sentence to collect manual evaluations of various types of correction sentences.
The following four representative GEC systems were used.

    \paragraph{SMT:} The model is based on statistical machine translation (SMT), a method that learns the translation probability of each word or phrase and the probability of the correct sequence as statistical information.
    Moses \cite{DBLP:conf/acl/KoehnHBCFBCSMZDBCH07} was used as a tool for the SMT.
    KenLM \cite{heafield-2011-kenlm} was used to train the language model.
    
    \paragraph{RNN:} A sequence-to-sequence transformation model based on recurrent neural network, which is a neural network method that takes into account information about the time series of data.
    The implementation is based on fairseq \cite{ott-etal-2019-fairseq}.
    Experiments were conducted using bi-directional LSTM.
    The number of word dimensions was set to 512, the batch size was set to 32, and the rest of the implementation settings followed those in \cite{DBLP:journals/corr/LuongPM15}.
    
    \paragraph{CNN:} A sequence-to-sequence transformation model based on convolutional neural network that learns by abstracting features of the data.
    The implementation is based on fairseq.
    The dimensions of the encoder and decoder were set to 512, and the remainder of the implementation settings were followed \cite{Chollampatt_Ng_2018}.
    
    \paragraph{Transformer:} A sequence-to-sequence transformation model consisting only of a mechanism called attention, which represents the attention of words in a sentence.
    The implementation is based on fairseq.
    The parameter settings followed those of \cite{NIPS2017_3f5ee243}.

\subsection{Datasets}
\label{Japanese Learner Corpus}
To obtain pairs of error-containing and corrected sentences, we used the TEC-JL \citelanguageresource{koyama-etal-2020-construction}. and FLUTEC \footnote{\scriptsize	{\url{https://github.com/kiyama-hajime/FLUTEC}}} as the data set with error-containing sentences.

\begin{table}[t]
  \label{table:data_statistics}
  \centering
  \begin{tabular}{lrr}
    \toprule
    Corpus & Essays & Sentence pairs \\
    \midrule
    Lang-8 & 192,673 & 1,296,114 \\
    TEC-JL & 139 & 2,042 \\
    FLUTEC & 169 & 2,100 \\
    \bottomrule
  \end{tabular}
    \caption{Number of essays and number of sentence pairs in the dataset used in the experiment. For Lang-8, it reports the number of essays and sentence pairs in Japanese.}
\end{table}

\paragraph{TEC-JL.}
TEC-JL includes 139 essays from Lang-8 containing 2,042 sentence pairs between learners' sentences and corrections.
The corrections were carried out by Japanese native speakers with minimal edits.
Unlike the original Lang-8 corpus, the TEC-JL is a relatively reliable dataset for evaluation.
In our experiments, we selected this dataset for comparing the performances of our quality estimation model with the original automatic evaluation methods using reference sentences.

\paragraph{FLUTEC.}
This dataset was annotated with sentences from Japanese language learners sampled from the Lang-8 corpus to generate multiple fluency-aware corrective sentences.
The dataset consists of dev and test data, with 1,050 sentences sampled each.
Note that in the experiment in Sec. \ref{sec:experiments}, we used only the dataset created from TEC-JL.

\subsection{Annotation}
\label{Annotation}
\paragraph{Policies.}
We obtained a set of pairs of Japanese learner sentences and corrected sentences by having four GEC systems correct the sentences of Japanese learners in the TEC-JL and FLUTEC.
To create an effective dataset for building a quality estimation model, we conducted annotation based on two policies:
1) Holistic evaluation. 
2) Annotation based on pairs of Japanese learners and corrected sentences.

First, \newcite{yoshimura-etal-2020-reference} collected human ratings of corrected sentences on three measures of grammaticality, fluency, and meaning preservation, and fine-tuned BERT on each measure to build a quality estimation model.
As a result, the model trained on grammaticality showed the best correlation with human ratings compared to reference-based automatic evaluation methods.
Borrowing from this, in order to create a quality estimation dataset at a low cost, we collected human ratings using a single grammaticality-oriented holistic rating scale.

Second, in \newcite{yoshimura-etal-2020-reference}, grammaticality was evaluated by examining only the corrected sentence using a simple five-step evaluation rule was used.
In the present study, to consider meaning preservation on a single evaluation scale, the original and corrected sentences were evaluated in pairs, and the rules were designed based on the evaluation of both before and after correction.

\paragraph{Process.} Four GEC systems were used to generate corrected sentences for 2,042 sentences written by Japanese learners included in TEC-JL. 
We asked three native Japanese university students to evaluate 4,391 sentence pairs, excluding the duplicated sentence pairs.
For a rules agreement among the annotators, the evaluations of up to 2,000 sentences were discussed.
The cases where scores differed by more than two were discussed among the annotators, and the guidelines were updated to supplement the rules for cases where evaluations were likely to differ.

\begin{table*}[t]
    \centering
    \begin{tabular}{cp{14cm}}
    \toprule
    Score & Description \\
    \midrule
    4     & S2 is grammatically correct or all significant and minor errors in S1 have been corrected.\\
    3     & All significant errors in S2 have been corrected. Some minor errors are acceptable. (e.g., Sentences containing not wrong, but unnecessary edits. Sentences containing spaces.)\\
    2     & S2 is able to correct important errors, but unacceptable minor errors are not corrected. Minor erroneous corrections have been made. (e.g., Missing punctuation.
Sentences ending with a reading mark or comma. N.B. These errors can have a score of 3 at annotator's decision.)\\
    1     & S2 is not able to correct significant and minor errors. Serious incorrect corrections have been made. (e.g., Changes to nouns and verbs that significantly impair the meaning of the original text.
Cases where the meaning has changed significantly from the original text even if they are grammatically correct.)\\
    0     & S1 is a sentence that is difficult to correct. Sentence that cannot be corrected. (e.g., Sentences where more than half of the text is non-Japanese)\\
    \bottomrule
    \end{tabular}
    \caption{Description of evaluation scores. S1 is the original text and S2 is the corrected text.}
    \label{tab:evaluation_scores}
\end{table*}

\paragraph{Guidelines.}
We used a 5-point Likert scale scheme to annotate grammaticality-based evaluation scores.
The evaluation scores were determined according to the rules described in Table \ref{tab:evaluation_scores}.~\footnote{In addition, we did not consider the following usages. First, the conversion from present tense to past tense is not treated as an error. Second, commas and periods were treated the same as reading and punctuation marks. Third, we ignore emoticons and symbols regardless of their position.}

In the experiment, the average ratings of the three annotators were used as the final manual evaluation. 
As an exception, in order to correctly reflect the annotator's evaluation, cases where zero was included in the three evaluations were handled as follows: cases when one of the three annotators rated the case as 0, the average of the other two annotators' ratings was taken; when two or more annotators rated the case as 0, no average was taken but the rating was set to 0.

\subsection{Analysis}

\begin{table*}[t]
  \centering
  \small
  \begin{tabular}{llc}
    \toprule
    Source text & Corrected sentence & Scores (avg.) \\
    \midrule
    \begin{CJK}{UTF8}{ipxm}このアルバイトを本当に楽しみ\end{CJK}\begin{CJK}{UTF8}{ipxg}にします\end{CJK}\begin{CJK}{UTF8}{ipxm}。\end{CJK} & \begin{CJK}{UTF8}{ipxm}このアルバイトを本当に楽しみ\begin{CJK}{UTF8}{ipxg}です\end{CJK}\begin{CJK}{UTF8}{ipxm}。\end{CJK}\end{CJK} & 2, 3, 3 (2.67) \\
    (I \textit{will} really look forward to this part-time job.) & (I \textit{am} really looking forward to this part-time job.) & \\
    \midrule
    \begin{CJK}{UTF8}{ipxm}元々沖縄へ\end{CJK}\begin{CJK}{UTF8}{ipxg}行きたいでした\end{CJK}\begin{CJK}{UTF8}{ipxm}。\end{CJK} & \begin{CJK}{UTF8}{ipxm}元々沖縄へ\end{CJK}\begin{CJK}{UTF8}{ipxg}行きたかったです\end{CJK}\begin{CJK}{UTF8}{ipxm}。\end{CJK} & 4, 4, 4 (4.00) \\
    (Originally, I \textit{want to went} to Okinawa.) & (Originally, I \textit{wanted to go} to Okinawa.) & \\
    \bottomrule
  \end{tabular}
    \caption{Annotation examples. The numbers in the Scores column represent the scores of the three annotators.}
  \label{table:data_example}
\end{table*}

\begin{figure}[t]
\begin{center}
\includegraphics[scale=0.4]{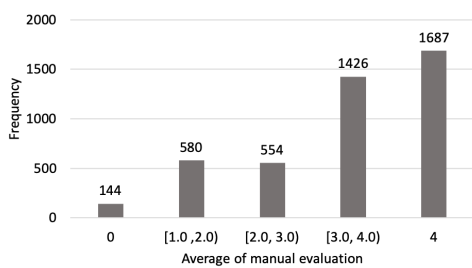}
\caption{Histogram of manual evaluation scores for sentence pairs generated from TEC-JL.}
\label{fig:Hist-TECJL}
\end{center}
\end{figure}

\begin{figure}[t]
\begin{center}
\includegraphics[scale=0.4]{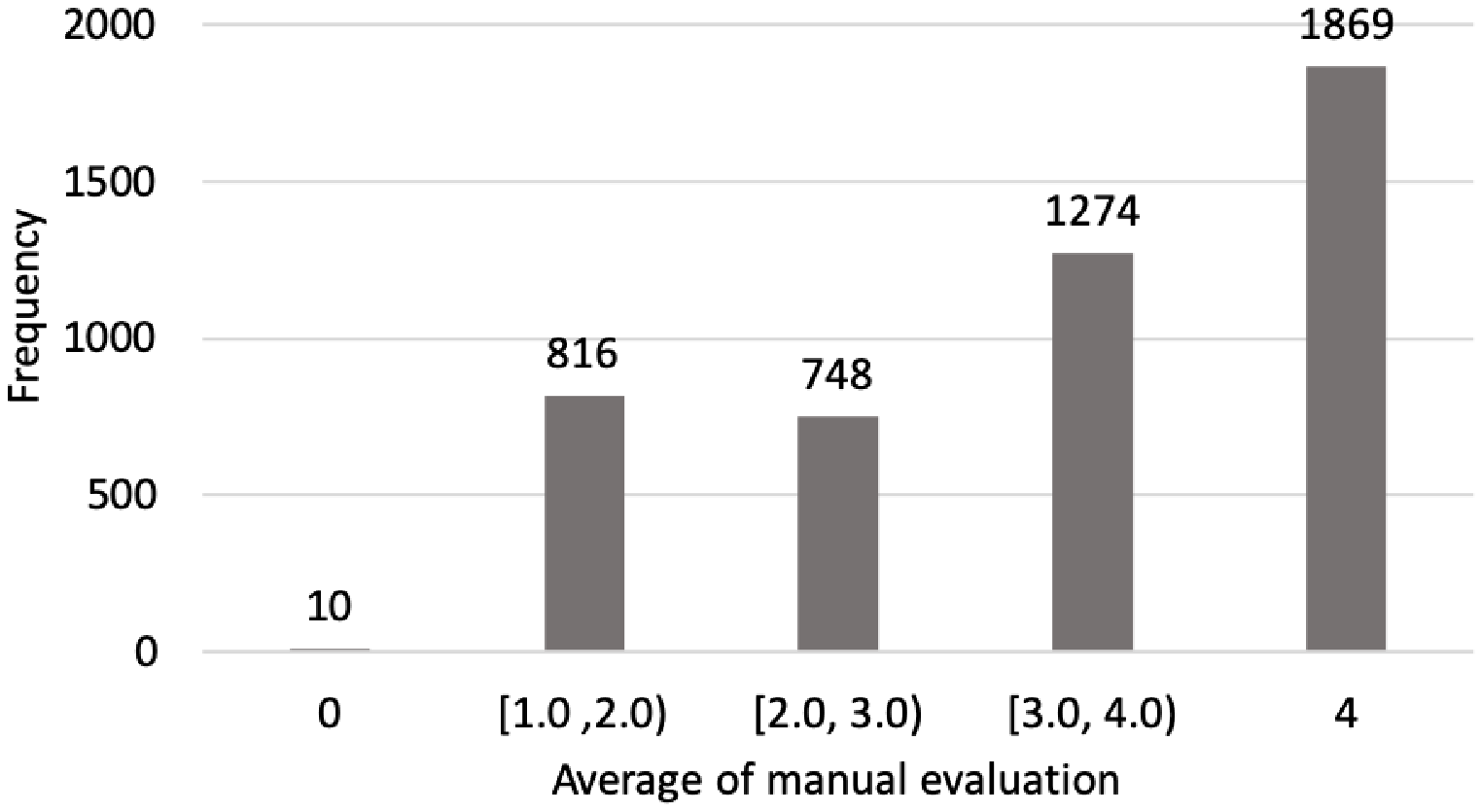}
\caption{Histogram of manual evaluation scores for sentence pairs generated from FLUTEC.}
\label{fig:Hist-FLUTEC}
\end{center}
\end{figure}

To measure the agreement rate between annotators, we used the kappa coefficient.
The value was found to be 0.49, indicating moderate agreement \cite{Landis77}.
Figures \ref{fig:Hist-TECJL} and \ref{fig:Hist-FLUTEC} show histograms of manual evaluation scores for sentence pairs created from TEC-JL and FLUTEC.\footnote{Because of the way averages are taken, there is no rating greater than 0 and less than 1.}
For both datasets, overall number of ratings is 3 to 4.
In the graph for TEC-JL, the small number of 0 ratings is probably due to the fact that low noise cases were selected by random sampling; for FLUTEC, the number of 0 ratings is very small due to the fact that low noise data was manually selected during sampling.

Table \ref{table:data_example} shows actual annotation examples.
In the above annotation example, a wrong sentence was corrected only partially.
The annotator evaluated this failure as a minor error, and rated the correction 2 or 3 at the annotator's discretion.
In the annotation example below, all annotators gave a rating of 4 because the valid and sufficient corrections were made.

\section{QE Experiments}
\label{sec:experiments}

To evaluate the quality estimation performance of the Japanese GEC, we measured the sentence-level correlation coefficients between the human evaluation scores of the reference-based and reference-less evaluations for the output of the GEC system.

\subsection{Settings}
To build a quality estimation model, we used the same method as \newcite{yoshimura-etal-2020-reference} to fine-tune BERT.
The input of BERT is the language learner's written and corrected sentences, and the output is the evaluation score.
Because \newcite{yoshimura-etal-2020-reference} assessed grammaticality, fluency, and meaning preservation separately, only the corrected sentences were used as input to the quality estimation model for grammaticality.
However, in this study, because we also assessed meaning preservation at once, both the learner's sentences and the corrected sentences were used as input for BERT.
For the output, we changed the output layer of the BERT to a regression model to output the ratings.

As for the proposed method, we performed a 10-fold cross-validation using the dataset created in Section \ref{Annotation}.
We divided the dataset into 10 parts so that the ratio of training, development, and testing was 8:1:1 and fine-tuned BERT with the training data.
Using dev data, we selected BERT hyperparameters with maximum Pearson's correlation coefficient by grid search with maximum sentence length 128, 256, batch size 8, 16, 32, learning rate 2e-5, 3e-5, 5e-5, and number of epochs 1-10.

For the baseline, we used GLEU as a reference-based method.
Because GLEU is evaluated with a value between 0 and 1, the evaluation was multiplied by 4 for comparison.
As reference sentences, we used two sentences from TEC-JL.
For the test data, we calculated the GLEU scores, quality estimation scores, and measured sentence-level correlations with manual evaluations. For meta-evaluation, we used Pearson's correlation coefficient and Spearman's rank correlation coefficient between the scores of the automatic and the manual evaluations for each sentence in the test data.

\subsection{Results}
\begin{table}[t]
    \centering
    \begin{tabular}{crr}
    \toprule
    Method   & Pearson & Spearman \\
    \midrule
      GLEU   & 0.320 & 0.362 \\
      fine-tuned BERT   & \textbf{0.580} & \textbf{0.413} \\
    \bottomrule
    \end{tabular}
    \caption{Results of meta-evaluation.}
    \label{table:results}
\end{table}

\begin{figure}[t]
\begin{center}
\includegraphics[scale=0.21]{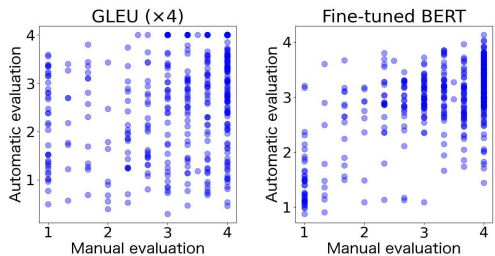}
\caption{Sentence-level correlation between automatic and manual evaluation scores.}
\label{fig:distributionchart}
\end{center}
\end{figure}

Table \ref{table:results} shows the results of sentence-level comparison for each automatic and manual evaluation score using Pearson's correlation coefficient and Spearman's rank correlation coefficient.
The results of the experiment show that fine-tuned BERT is more highly correlated with human ratings than GLEU in both Pearson's correlation coefficient and Spearman's rank correlation coefficient.
Figure \ref{fig:distributionchart} shows that there is almost no correlation between the GLEU scores and the manual ratings, while the fine-tuned BERT model shows a weak correlation with the manual ratings, and the model tends to do a proper quality estimation  especially for the cases rated 1.

\subsection{Analysis}
\begin{table*}[t]
    \centering
    \begin{tabular}{lll}
    \toprule
        & Example 1 & Example 2 \\
    \midrule
        S & \begin{CJK}{UTF8}{ipxg}ソンあ\end{CJK}\begin{CJK}{UTF8}{ipxm}さんが\end{CJK}\begin{CJK}{UTF8}{ipxg}好きがっている\end{CJK}\begin{CJK}{UTF8}{ipxm}のは推理小説です。\end{CJK} & \begin{CJK}{UTF8}{ipxg}堤　真一\end{CJK}\begin{CJK}{UTF8}{ipxm}さんの演技もよかったですね。\end{CJK} \\
        & (\textit{Son A wants to like} reading mysteries.) & (\textit{Shinichi Tsutsumi's} acting was also good.) \\
    \midrule
        C &  \begin{CJK}{UTF8}{ipxg}ソンあ\end{CJK}\begin{CJK}{UTF8}{ipxm}さんが\end{CJK}\begin{CJK}{UTF8}{ipxg}好きになっている\end{CJK}\begin{CJK}{UTF8}{ipxm}のは推理小説です。\end{CJK} & \begin{CJK}{UTF8}{ipxg}写真一\end{CJK}\begin{CJK}{UTF8}{ipxm}さんの演技もよかったですね。\end{CJK} \\
        & (\textit{Son A is becoming fond} of mystery novels.) & (\textit{Photo 1's} performance was also good.) \\
    \midrule
        R & \begin{CJK}{UTF8}{ipxg}ソンア\end{CJK}\begin{CJK}{UTF8}{ipxm}さんが\end{CJK}\begin{CJK}{UTF8}{ipxg}好んでいる\end{CJK}\begin{CJK}{UTF8}{ipxm}のは推理小説です。\end{CJK} & \begin{CJK}{UTF8}{ipxg}堤　真一\end{CJK}\begin{CJK}{UTF8}{ipxm}さんの演技もよかったですね。\end{CJK} \\
        & (\textit{Sonna likes} to read mysteries.) & (\textit{Shinichi Tsutsumi's} acting was also good.) \\
        & \begin{CJK}{UTF8}{ipxg}ソンア\end{CJK}\begin{CJK}{UTF8}{ipxm}さんが\end{CJK}\begin{CJK}{UTF8}{ipxg}好きな\end{CJK}\begin{CJK}{UTF8}{ipxm}のは推理小説です。\end{CJK} & \begin{CJK}{UTF8}{ipxg}堤　真一\end{CJK}\begin{CJK}{UTF8}{ipxm}さんの演技もよかったですね。\end{CJK} \\
        & (\textit{Sonna likes} to read mysteries.) & (\textit{Shinichi Tsutsumi's} acting was also good.) \\
    \midrule
        E & GLEU, BERT, Human / 0.59, 3.30, 3.00 & GLEU, BERT, Human / 3.14, 1.12, 1.00 \\
    \bottomrule
    \end{tabular}
    \caption{Successful examples of the quality estimation method. The first line shows the source text (S), the second line shows the corrected sentence (C), the third line shows the two reference sentences (R), and the fourth line shows automatic and manual evaluations (E).}
    \label{table:successful}
    \vspace{0.5cm}
    \centering
    \begin{tabular}{lll}
    \toprule
        & Example 1 & Example 2 \\
    \midrule
        S & \begin{CJK}{UTF8}{ipxm}社会人になりたくない\end{CJK}\begin{CJK}{UTF8}{ipxg}原因\end{CJK}\begin{CJK}{UTF8}{ipxm}が多い\end{CJK}\begin{CJK}{UTF8}{ipxg}だ\end{CJK}\begin{CJK}{UTF8}{ipxm}。\end{CJK} & \begin{CJK}{UTF8}{ipxg}ほんと\end{CJK}\begin{CJK}{UTF8}{ipxm}ですか？\end{CJK} \\
        & (There are many \textit{cause} why I don't want to be a working adult.) & (\textit{Really}?) \\
    \midrule
        C &  \begin{CJK}{UTF8}{ipxm}社会人になりたくない\end{CJK}\begin{CJK}{UTF8}{ipxg}理由\end{CJK}\begin{CJK}{UTF8}{ipxm}が多い。\end{CJK} & \begin{CJK}{UTF8}{ipxg}本当\end{CJK}\begin{CJK}{UTF8}{ipxm}ですか？\end{CJK} \\
        & (There are many \textit{reasons} why I don't want to be a working adult.) & (\textit{Really}?) \\
    \midrule
        R & \begin{CJK}{UTF8}{ipxm}社会人になりたくない\end{CJK}\begin{CJK}{UTF8}{ipxg}原因\end{CJK}\begin{CJK}{UTF8}{ipxm}が多い。\end{CJK} & \begin{CJK}{UTF8}{ipxg}本当\end{CJK}\begin{CJK}{UTF8}{ipxm}ですか？\end{CJK} \\
        & (There are many \textit{cause} why I don't want to be a working adult.) & (Really?) \\
        & \begin{CJK}{UTF8}{ipxm}社会人になりたくない\end{CJK}\begin{CJK}{UTF8}{ipxg}原因\end{CJK}\begin{CJK}{UTF8}{ipxm}が多い。\end{CJK} & \begin{CJK}{UTF8}{ipxg}ほんと\end{CJK}\begin{CJK}{UTF8}{ipxm}ですか？\end{CJK} \\
        & (There are many \textit{cause} why I don't want to be a 
    working adult.) & (\textit{Really}?) \\
    \midrule
        E & GLEU, BERT, Human / 2.63, 1.45, 4.00 & GLEU, BERT, Human / 3.41, 2.03, 4.00 \\
    \bottomrule
    \end{tabular}
    \caption{Failed examples of the quality estimation method. The first line shows the source text (S), the second line shows the corrected sentence (C), the third line shows the two Reference sentences (R), and the fourth line shows automatic and manual evaluations (E).}
    \label{table:failed}
\end{table*}

To investigate the differences in the evaluations, we analyzed an example of evaluation by the two automatic evaluation methods: the quality estimation BERT model and the reference-based GLEU.

\paragraph{Successful cases.}
Table \ref{table:successful} shows two examples where the quality estimation model was able to evaluate correctly.

The corrected sentence in Example 1 is grammatically correct, but the word \begin{CJK}{UTF8}{ipxm}``ソンあ (Sonna)''\end{CJK}, which is supposed to be a person's name, is expressed in a mixture of \textit{katakana} and \textit{hiragana}, resulting in a manual evaluation of 3.0\footnote{In Japanese, katakana should be used for transliteration.}.
The reference sentence has two types of correction for the expression \begin{CJK}{UTF8}{ipxm}``好きがっている (wants to like)''\end{CJK}, however, because this is not an expression that can be corrected by the GEC system, its evaluation widely differs from the manual evaluation.
However, the quality estimation model was relatively close to the manual evaluation.

In Example 2, the person's name \begin{CJK}{UTF8}{ipxm}``堤　真一 (Tsutsumi Shinnichi)''\end{CJK} is corrected to \begin{CJK}{UTF8}{ipxm}``写真一 (Photo 1)''\end{CJK} in the corrected text, and the manual evaluation is 1 because the meaning of the original text is greatly impaired by this correction.
In the reference sentence, no correction was made to the source text, and the output of the GEC system was superficially similar to the source text.
Because GLEU calculates the score by subtracting the number of n-grams that appear in the source text but do not appear in the reference text from the number of n-gram matches between the corrected text and the reference text, it gives a high score to the corrected text.
Meanwhile, the quality estimation model captures the meaning changes between the source and reference sentences and can provide an evaluation similar to a manual evaluation.

\paragraph{Failed cases.}
We analyzed examples of evaluations in which the quality estimation model could not be correctly evaluated.
Table \ref{table:failed} shows two examples where the quality estimation model failed to evaluate correctly.

In Example 1, the manual evaluation is 4.0 because the correct correction is made but the quality estimation model's evaluation widely differs from from the manual evaluation.
Meanwhile, GLEU can determine from the reference sentences that the deletion of the \begin{CJK}{UTF8}{ipxm}``だ (copula)''\end{CJK} at the end of the word is a correct correction, and thus it is closer to the manual evaluation than the quality estimation model.

In Example 2, the quality estimation model does not recognize the edit of \begin{CJK}{UTF8}{ipxm}``ほんと (really)''\end{CJK} to \begin{CJK}{UTF8}{ipxm}``本当 (really)''\end{CJK} as a correct correction\footnote{\begin{CJK}{UTF8}{ipxm}``本当''\end{CJK} is pronounced as \begin{CJK}{UTF8}{ipxm}``ほんとう''\end{CJK} and \begin{CJK}{UTF8}{ipxm}``ほんと''\end{CJK} is a colloquial expression of \begin{CJK}{UTF8}{ipxm}``本当''\end{CJK}.}, while GLEU can recognize the edit as a correct correction from the reference sentence.
By contrast, GLEU recognizes that the edit is a correct correction from the reference sentence.

\section{Conclusions}
In this study, we constructed a dataset that includes a manual evaluation of a grammaticality-oriented holistic approach added to the outputs of Japanese GEC systems.
This dataset consisted of three elements: source text, corrected text, and human ratings.
The source text consisted of the Lang-8 corpus, and the corrected text consisted of the outputs of the four GEC systems.
The human ratings are annotated by students whose first language is Japanese.

Using the dataset, we optimized BERT directly on human ratings to create a quality estimation model.
To compare the performance of the reference-less and reference-based methods, we measured the sentence-level correlation coefficients between the evaluation scores of each method and the human evaluation scores for the output of the GEC systems.
The experimental results showed that the quality estimation model had a higher correlation with manual evaluation than the method using reference sentences, demonstrating the usefulness of a reference-less automatic evaluation method for Japanese GEC.

Future developments include a more detailed analysis using error-type annotation and research to measure the GEC performance by selecting the output with the highest QE score from the outputs of multiple GEC systems by reranking.

\section{Bibliographical References}\label{reference}

\bibliographystyle{lrec2022-bib}
\bibliography{lrec2022-jaeval}

\section{Language Resource References}
\bibliographystylelanguageresource{lrec2022-bib}
\bibliographylanguageresource{languageresource}

\end{document}